\documentclass{article}
\usepackage{spconf,amsmath,graphicx}

\usepackage{enumitem}
\usepackage{subcaption}
\usepackage{amsmath}
\usepackage{amssymb}
\setlist{nosep, leftmargin=14pt}

\title{Multiscale Neuroimaging Features for the Identification of Medication Class and Non-Responders in Mood Disorder Treatment}
\name{Bradley T. Baker$^{\star \dagger \ddagger}$\qquad Mustafa S. Salman$^{\star}$\qquad Zening Fu$^{\star}$\qquad Armin Iraji$^{\star\dagger}$
\qquad
Elizabeth Osuch$^{\diamond \spadesuit}$\qquad H. Jeremy Bockholt$^{\star \dagger}$ Vince D. Calhoun$^{\star \dagger \ddagger \blacksquare}$ \qquad}
\address{${}^\star$Tri-Institutional Center for Translational Research in Neuroimaging and Data Science (TReNDS)\\\{$^\dagger$Georgia State University, $^\ddagger$Georgia Institute of Technology, ${}^\blacksquare$Emory University\}Atlanta, GA, USA\\${}^\diamond$    Lawson Health Research Institute, London Health Sciences Centre, FEMAP, London, Ontario, Canada\\${}^\spadesuit$    Schulich School of Medicine and Dentistry, Western University, London, Ontario, Canada}
\begin{document}

%
\maketitle
%
\begin{abstract}
In the clinical treatment of mood disorders, the complex behavioral symptoms presented by patients and variability of patient response to particular medication classes can create difficulties in providing fast and reliable treatment when standard diagnostic and prescription methods are used. Increasingly, the incorporation of physiological information such as neuroimaging scans and derivatives into the clinical process promises to alleviate some of the uncertainty surrounding this process. Particularly, if neural features can help to identify patients who may not respond to standard courses of anti-depressants or mood stabilizers, clinicians may elect to avoid lengthy and side-effect-laden treatments and seek out a different, more effective course that might otherwise not have been under consideration.  Previously, approaches for the derivation of relevant neuroimaging features work at only one scale in the data, potentially limiting the depth of information available for  clinical decision support. In this work, we show that the utilization of multi spatial scale neuroimaging features - particularly resting state functional networks and functional network connectivity measures - provide a rich and robust basis for the identification of relevant medication class and non-responders in the treatment of mood disorders.  We demonstrate that the generated features, along with a novel approach for fast and automated feature selection, can support high accuracy rates in the identification of medication class and non-responders as well as the identification of novel, multi-scale biomarkers.
\end{abstract}

\section{Introduction}
\label{sec:intro}

Numerous studies have reported underlying neural differences between individuals exhibiting different classes of mood disorder such as major depressive disorder (MDD), bipolar disorder (BPD) and others \cite{ de2013distinguishing,  kempton2011structural, osuch2018complexity}. In clinical settings, the subtle phenotypical distinctions between unipolar and bipolar disorders can often complicate diagnosis and treatment. Thus, the process of finding the correct medication class for an individual can involve long periods of ineffective trial periods. If the choice of medication is incorrect due to initial misdiagnosis, the patient's symptoms may not be adequately addressed by a correspondingly incorrect medication while still being subject to detrimental side effects of that medication. Even when the psychiatric diagnosis is correct, the effectiveness of particular medication classes may still vary between individuals \cite{ball2014toward, trivedi2016establishing, davis2019white}. Increasingly, there is a desire in the research and clinical community to discover underlying patient biomarkers that may help predict patient response to particular medication classes \cite{ball2014toward, trivedi2016establishing, osuch2018complexity, davis2019white, salman2023multi}, with the ultimate aim being to provide clinicians with models of medication response as a clinical support tool.

Functional magnetic resonance imaging (fMRI) is a neuroimaging modality in which an endogenous blood oxygenation level dependence (BOLD) contrast is measured at multiple time points in a scanning session \cite{ogawa1990brain}. Biomarkers extracted from fMRI have shown to be useful in distinguishing between MDD and BPD \cite{he2019altered, han2020resting, rai2021default} as well as responsiveness to particular medication classes \cite{trivedi2016establishing, osuch2018complexity, salman2023multi}. Particularly, intrinsic connectivity networks (ICNs) extracted from resting-state fMRI, such as the striatum precuneus \cite{he2019altered}, default mode and fronto-parietal networks \cite{rai2021default} provide basis for further targeted studies and treatment involving the role of these networks in mood disorders and medication response. 


A previous multi-study work approached the prediction of medication response \cite{salman2023multi} by performing spatially constrained independent component analysis (scICA) with the Neuromark\_fMRI\_1.0\ template \cite{du2020neuromark}.  Out of the 53 large-scale ICNs from Neuromark, 7 domain-specific networks were selected via sequential forward selection (SFS) and applied to a kernel support vector machine (SVM)  learning model for the prediction of whether patients should receive antidepressants (AD) or mood stabilizers (MS). Across three studies \cite{osuch2018complexity, trivedi2016establishing, poldrack2016phenome}, the kernel SVM provided around 90\% accuracy when the non-responder population was excluded.

Recently, a new Neuromark template was introduced which expands on the previous framework by incorporating ICNs extracted across multiple spatial resolutions \cite{iraji2023identifying}. In this work, we demonstrate that ICNs from this multi-scale template provide useful biomarkers for the prediction of medication response. Particularly, we improve on the analysis in \cite{salman2023multi} by showing that the features from the multi-scale template provide a reliable prediction of patient medication response, while also aiding in the prediction of non-responsiveness to medication. Furthermore, we illustrate that a soft sequential feature selection approach allows for an improvement over the original ICNs selected in \cite{salman2023multi}. We demonstrate our approach on data collected from Western University \cite{osuch2018complexity}, predicting whether or not the patient will respond to medication, as well as predicting which medication class subsequently produced the best response.
\section{Methods}
\label{sec:methods}


\subsection{Data set}
\label{sec:methods-data}

Our primary data set in this work is an fMRI study collected from Western University \cite{osuch2018complexity, salman2023multi}. The subjects were between 16 and 27 years of age, with no significant effect of age between groups (p = 0.1492). The study includes 147 subjects, divided into four groups: 33 controls, 35 patients with BD type-I, 67 with MDD, and 12 with unknown diagnoses.  Diagnoses were performed using the structural clinical interview for DSM disorders-IV (SCID-IV) or the diagnostic interview for genetic studies (DIGS), and  were confirmed by clinical psychiatric diagnostic assessment. Agreement between SCID-IV/DIGS diagnosis and clinical diagnosis was required for the patients. If there was disagreement between DIGS and clinical diagnosis or if patients had one or more first-degree relatives with mental illness, they were categorized as the ”unknown” group.  The study clinician determined medication class using a chart review to treat each patient to attain sustained euthymia, lasting at least six months. Medication-class was simplified to either an AD or MS (lithium, lamotrigine, carbamazepine, divalproex sodium). Based on medication class, we divided the 147 subjects into four groups: 33 controls, 47 patients responding to AD, 45 responding to MS, 8 nonresponders, and 14 remitted without medication.

MRI data were obtained at the Lawson Health Research Institute using a 3.0T Siemens Verio MRI scanner and a 32-channel phased-array head coil. The data include gradient-echo, echo-planar imaging (EPI) scans with the following acquisition parameters: repetition time (TR) = 2000 ms, echo time (TE) = 30 ms, 40 axial slices and thickness = 3 mm, with no parallel acceleration, flip angle = 90°, field of view (FOV) = 240 × 240 mm, matrix size = 80 × 80. The length of the resting fMRI scan was approximately 8 min, and 164 brain volumes were collected.  Data were preprocessed and quality controlled using the statistical parametric mapping (SPM) software. We utilized the same preprocessing steps as in \cite{salman2023multi}, and refer the reader to that publication for details. 

\subsection{Feature Generation}
\label{sec:methods-features}
\subsubsection{Multi-Scale Template}

To extract ICNs from our resting-state data, we utilized a new multi-scale Neuromark template (Neuromark\_fMRI\_2.1\_modelorder-multi.nii, http://trendscenter.org/data)
 which contains 105 networks extracted across multiple spatial resolutions \cite{iraji2023identifying}. These ICNs were extracted from over 20 different studies and utilized an approach for group multi-scale ICA \cite{iraji2022multi} over 8 distinct model orders to extract components at different spatial resolutions. Out of 900 extracted ICNs, components were then manually selected based on commonly accepted criteria for ICN such as peak values in gray matter, low spatial overlap with vascular and ventricular structures and low spatial similarity with known artifacts \cite{iraji2023identifying}. A subset of components with similarity below 0.8 were retained leaving 105 remaining components in 6 functional domains: visual (VI), cerabellar (CB), temporal (TP), subcortical (SC), somatomotor (SM) and higher cognitive (HC). 

For the sake of comparing the multi-scale template with the original Neuromark template, we also included the 53 ICNs extracted for the Western University data, replicating the analysis in \cite{salman2019group, salman2023multi}. This reference set included 53 labeled and ordered components from seven distinct functional domains: subcortical (SC), auditory (AD), sensorimotor (SM), visual (VI), executive control (CO), default mode (DM) and cerebellar (CB). 

\subsubsection{Spatially Constrained ICA}

To extract ICNs from our data, we used the spatially-constrained ICA (scICA) algorithm available in the GIFT fMRI toolbox \cite{du2020neuromark}. Furthermore, we estimated the functional network connectivity (FNC) matrix for each subject using the Pearson correlation coefficient between the time courses of 105 components, the dimension of which was 105 × 105. For comparison, we performed the same scICA+FNC analysis using the 53 original ICNs from the neuromark template as well. Following the precedent in \cite{salman2023multi} we  use computed spatial maps and FNC matrices as features for prediction.

\subsection{Medication-Class Identification}
\label{sec:methods-class}
\subsubsection{Kernel SVM}

Support vector machines (SVMs) are a class of machine learning model widely utilized for classification tasks. At their core, SVMs utilize a maximal margin linear classifier to create a decision boundary between high-dimensional feature spaces. In more complex decision tasks, a choice of nonlinear kernel function for the SVM allows for the creation of a nonlinear decision boundary. Following the precedent in \cite{salman2023multi}, we utilize a Riemannian kernel function based on the (cosine of the) principal angle between subspaces (PABS) distance metric. 
Formally, if we let $\mathbf{U}$ and $\mathbf{V}$ be the set of spatial maps in Voxel by Component space (i.e. $\mathbf{U},\mathbf{V}\in\mathbb{R}^{V \times K}$), then the angles in the PABS metric correspond to the ordered singular values of the matrix product $\mathbf{U}^\top \mathbf{F}$. If we define the PABS mapping as $S(\mathbf{U},\mathbf{V}) = \sum_{k}^{K} \sigma_k$, then the complete nonlinear kernel is computed as $K(\mathbf{U},\mathbf{V}) = \tanh(\gamma S(\mathbf{U},\mathbf{V}))$, where $\gamma$ is a scaling hyper-parameter which we  set to $\gamma = 1$. For a study with $N$ subjects, we can compute the full $N\times N$ kernel matrix by computing this scalar value between the spatial maps or FNCs for each subject pair. 




\subsubsection{Soft Sequential Forward Selection}

In order to determine which networks are most informative for the identification of patient medication response, we follow the precedent in \cite{salman2023multi} and reduce the complexity of our feature space by performing sequential forward selection to determine which ICNs best predict medication response.  In previous work, salient networks were identified by performing sequential forward selection, in which ICNs from each domain were included one by one, and the component providing the highest accuracy model was selected for the final set of features.  Because we are not simply performing binary classification, we use the area under the precision-recall curve in lieu of accuracy for our model selection criterion. Additionally, we perform a more exhaustive search by taking the top 5 performing feature sets during each stage the forward selection process, and only using the top-performing feature set after all domains have been searched. This soft selection approach allows us to include feature combinations which may have been otherwise excluded due to sub-par performance at earlier stages in the selection process, but which show overall better performance as more domains are included. We call this approach \textbf{soft sequential forward selection (SSFS),} to contrast it from the standard sequential forward selection.

\section{Results and Discussion}
\label{sec:results}

\begin{figure}[h!]
     \begin{subfigure}[b]{.49\linewidth}
    \includegraphics[width=\linewidth]{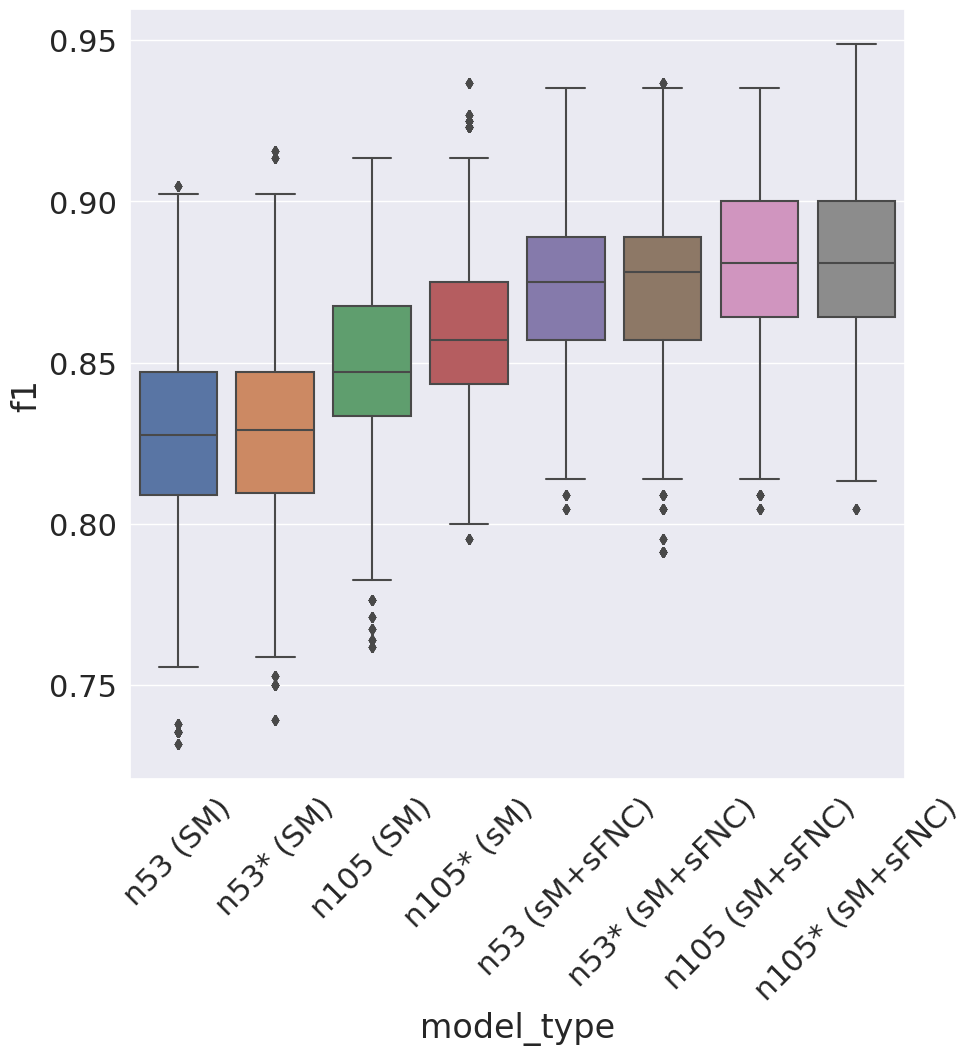}
    \caption{F1 Scores}
        \label{fig:model-performance-f1}
    \end{subfigure}
         \begin{subfigure}[b]{.5\linewidth}
    \includegraphics[width=\linewidth]{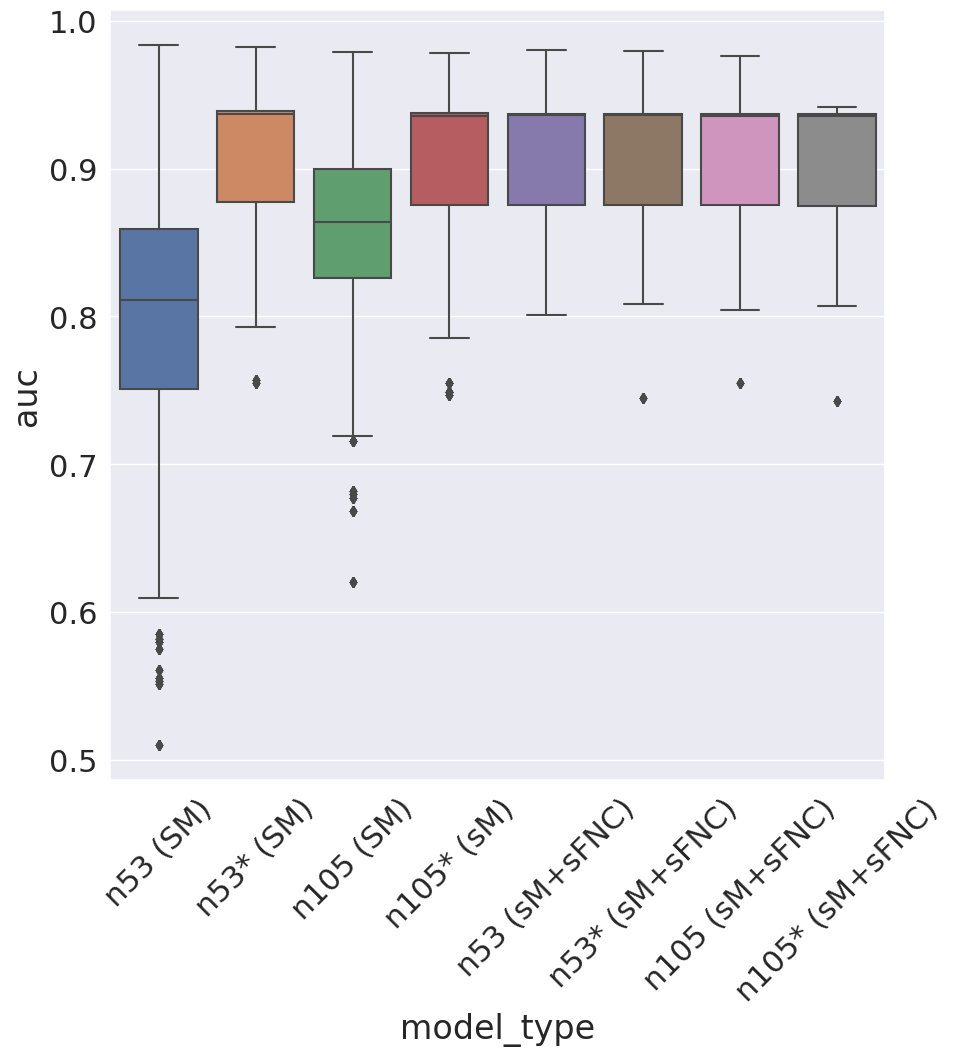}
    \caption{AUC Scores}
        \label{fig:model-performance-auc}
    \end{subfigure}
    \caption{Performance metrics of the Kernel SVM for different feature sets. We evaluated the original NeuroMark template (n53-Blue) and the best-matching components from the multiscale template (n105-Green). We then evaluated the performance of each template using SSFS (NeuroMark: n53*-Orange, MultiScale: n105*-Red). For each of these feature sets we evaluated the performance of utilizing spatial maps (SM) alone and spatial maps along with static FNC matrices (SM+sFNC). Performance metrics were aggregated over 1000 randomly initialized models and 5-fold cross validation.} 
    \label{fig:model-performance}
\end{figure}

In this section, we describe the results obtained on the task of medication-response identification using the multi-scale template in comparison with the neuromark single-scale template. Our main metric of performance is the area under the (precision-recall) curve (AUC) (see figure \ref{fig:model-performance-auc}), which provides more reliable evaluation of performance in multi-class settings than accuracy or precision by themselves. Additionally, we compute the F1 score (see figure \ref{fig:model-performance-f1}) and provide that as an additional evaluation metric, hoping to highlight which model classes perform best in this imbalanced scenario where relatively few subjects classify as non-responders. These computed metrics are shown in figure \ref{fig:model-performance} for both templates and different selections of functional features (Spatial Maps and FNCs) which we describe below.

\begin{figure}[h!]
    \begin{subfigure}[b]{0.49\linewidth}
    \includegraphics[width=\linewidth]{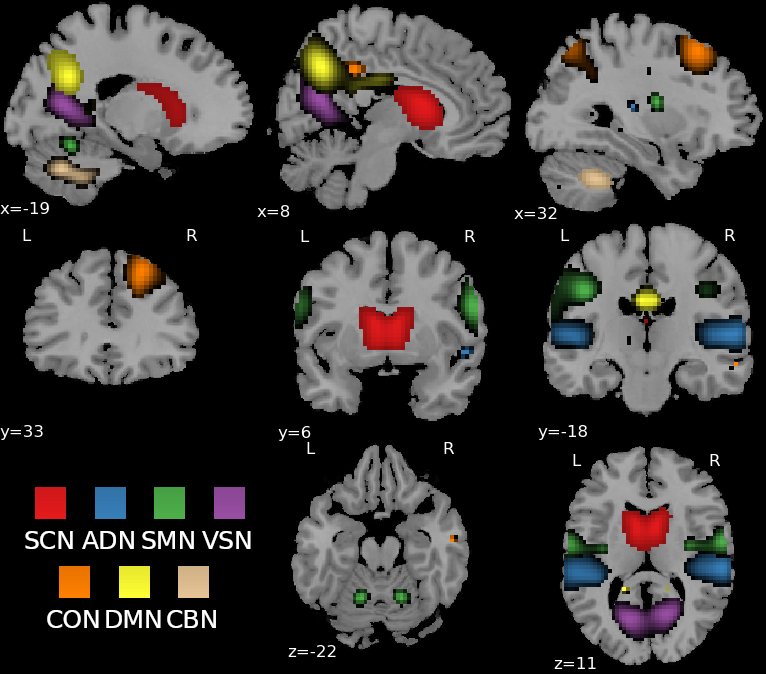}
    \caption{The Neuromark components previously selected in \cite{salman2023multi}.\vspace{0.2em}} 
    \label{fig:NM53-salman}
    \end{subfigure}
         \begin{subfigure}[b]{0.5\linewidth}
    \includegraphics[width=\linewidth]{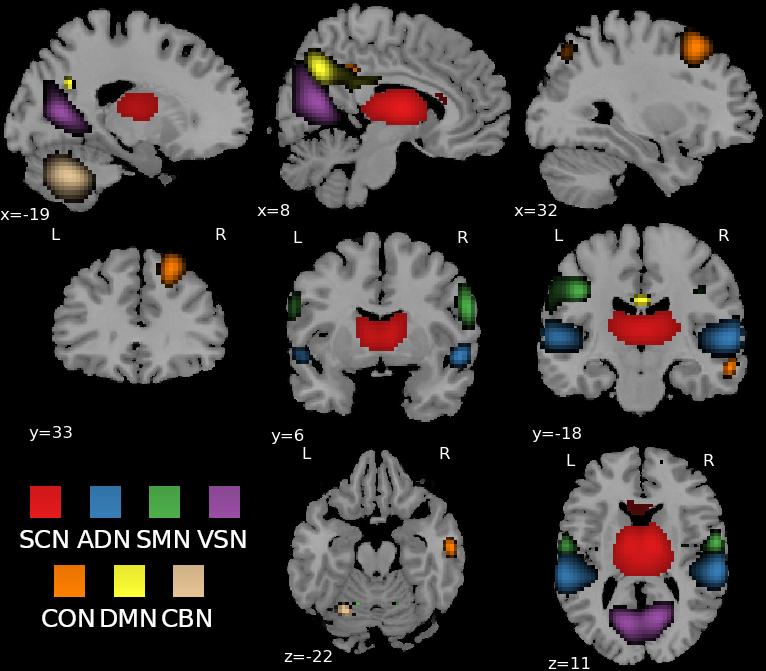}
        \caption{The multi-scale components which best match \cite{salman2023multi}}
        \label{fig:NM105-salman}        
    \end{subfigure}
    \caption{ICNs computed using sequential feature selection in  both templates. We visualize each ICN in different colors corresponding to Neuromark domains.}
    \label{fig:salman}
\end{figure}

In terms of the feature sets we choose to evaluate, we begin with the original 7 components selected from the neuromark template in \cite{salman2023multi}. These selected ICNs can be found in figure \ref{fig:NM53-salman}, and are identified as Sub-Cortical (Caudate), Auditory (Superior Temporal Gyrus), Somatomotor (Postcentral), Visual (Calcarine Gyrus), Control (Inferior Parietal Lobule), Default Mode (Precuneus), and Cerebellar. We treat these 7 components as our primary baseline set of features for comparison.  From the multi-scale template, we then also select the 7 components which are maximally spatially correlated with the original 7 neuromark ICNs. These networks can found in figure \ref{fig:NM105-salman} and can be seen to demonstrate significant overlap with the original 7 components plotted in figure \ref{fig:NM53-salman}.

\begin{figure}[h!]
     \begin{subfigure}[b]{0.49\linewidth}
    \includegraphics[width=\linewidth]{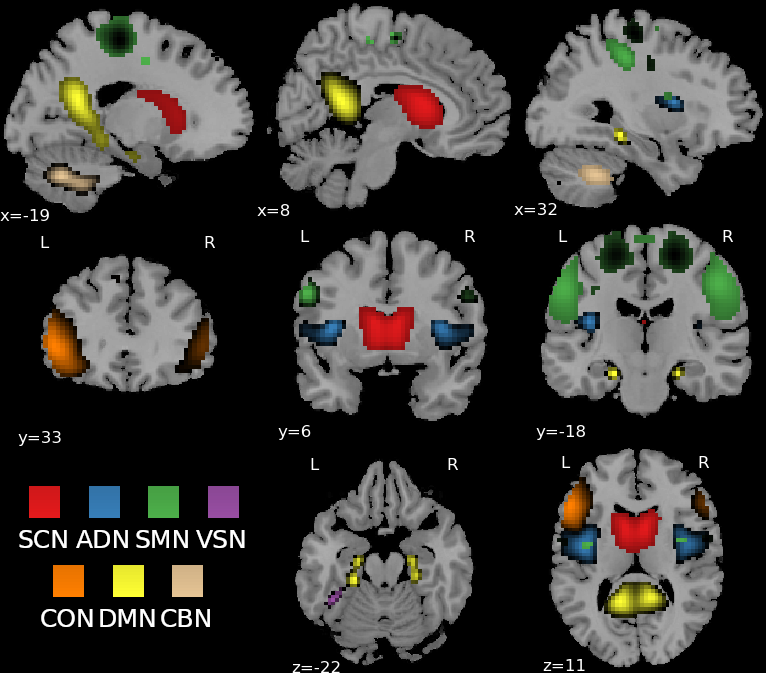}
    
    \caption{The ICNs selected from Neuromark using SSFS.\vspace{.1em}}
        \label{fig:NM53-sfss}
    \end{subfigure}
         \begin{subfigure}[b]{0.5\linewidth}
    \includegraphics[width=\linewidth]{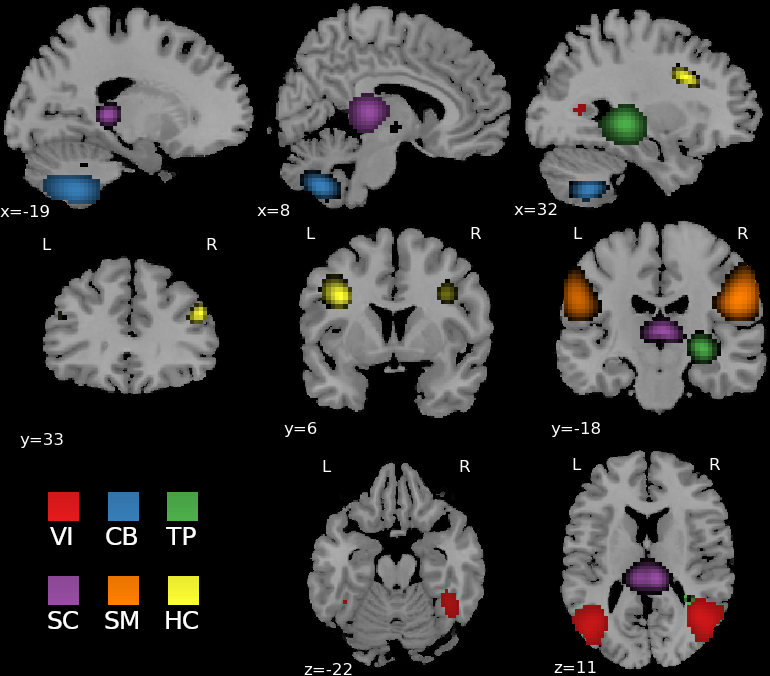}
    \caption{The ICNs selected from the Multi-scale template using SSFS.}
    \label{fig:N105-sfss}
    \end{subfigure}
    \caption{ICNs selected for each template using soft forward selection. Each of the ICNs  are visualized in different colors corresponding to the domains for the corresponding template.}
    \label{fig:sfss}
\end{figure}

To demonstrate the effectiveness of our soft sequential forward selection approach, we also compute distinct feature sets for both the original neuromark template (see figure \ref{fig:NM53-sfss}) and the multi-scale template (see figure \ref{fig:N105-sfss}). The 7 components selected for NeuroMark are identified as Sub-Cortical (Caudate), Auditory (Rolandic Operculum), Somatormotor (Postcentral Gyrus), Visual (Middle temporal gyrus), Control (Inferior Frontal Gyrus), Control (Precuneus) and Cerebellar.  For the multi-scale template, the 7 selected component correspond to the Visual, Cerebellar, Temporal, Subcortical, Somatomotor and Higher Cognitive domains.

For both templates and feature-selection types we evaluate the performance of the model using only spatial maps and then using both spatial maps and static FNC matrices. The AUC and F1 metrics can be found in figure \ref{fig:model-performance} for both template types, feature selection schema and spatial map and FNC inclusion. For each experiment, we perform 5-fold cross validation and train 1000 randomly-initialized models for each train-validation split. We then report the validation scores on holdout data over all of these individual models and  folds.

We can see from the performance of the final models in figure \ref{fig:model-performance} that even when using the originally matched components selected in \cite{salman2023multi}, the kernel SVM improves significantly when using the multi-scale spatial maps over the single scale maps (compare the blue and green boxes in figure \ref{fig:model-performance}). We see that the features computed using soft sequential forward selection provide a further performance boost over the original feature set utilized in \cite{salman2023multi}. Finally, we can see that incorporation of FNC features boost performance across all feature sets than when FNC features are not included. In terms of AUC performance, SFSS or inclusion of FNC features both bring both templates up to top performance; however, when we compare F1 scores, we can see that the multi-scale template outperforms the single-scale template even with SFSS and inclusion of FNC features. In terms of raw F1 score, the multi-scale template with SFSS utilizing both spatial maps and FNC features outperforms all model classes.

When we compare the features selected in \cite{salman2023multi} with those obtained via SFSS (see figures \ref{fig:salman} and \ref{fig:sfss}), we  notice some interesting differences between the ICNs which provide the best model performance for predicting medication response. For the original neuromark template, the Precuneus ICN in the Subcortical Domain and Cerebellar ICN appear in both selected feature sets, indicating that these networks are consistently predictive for medication response in the Western data set. In the other five domains, we see that SFSS provides slightly different ICN selections than in \cite{salman2023multi}, and though we clearly see from figure \ref{fig:model-performance} that these minor adjustments radically improve the overall AUC, the comparable F1 scores of these two feature sets indicate these differences may not be significant for this imbalance.

When we perform the SSFS process for the multi-scale template on its own, without any reference to the original neuromark template, the selected components are quite distinct from those found using the original neuromark template and from those selected in the multi-scale template by matching. We see some overlap between the Somatomotor components; however, the remaining domains identify distinct ICNs with differences such as the visual and subcortical components coming from further back in the brain and from a smaller model order, and the cerebellar ICN coming from lower in the cerebellum. Corresponding with the highest overall performance, these new multi-scale ICNs demand further neurological investigation and validation in further studies regarding response to medication class.

\section{Conclusion}
\label{sec:majhead}

In this work we utilized a new multi-scale template to generate novel features for the prediction of medication response for mood disorder treatment. We demonstrate that this multi-scale template, along with a more robust feature-selection schema allows for the computation of intrinsic connectivity networks (ICNs) which provide more robust performance for the identification of medication class than a single-scale reference template.  Following previous work, we utilized a kernel-SVM based classifier and demonstrated that even when non-responders are included, the ICNs provided from the multi-scale template along with a more flexible feature-selection schema allow for improved performance over the single scale case.


\bibliographystyle{IEEEbib}
\bibliography{strings,refs}

\end{document}